\documentclass[final]{cvpr}

\usepackage{times}
\usepackage{epsfig}
\usepackage{graphicx}
\usepackage{amsmath}
\usepackage{amssymb}
\usepackage{booktabs}
\usepackage{varioref}
\usepackage{placeins}
\usepackage{float}
\usepackage{tabularx}
\usepackage{enumitem}

\usepackage[pagebackref=true,breaklinks=true,colorlinks,bookmarks=false]{hyperref}

\def\cvprPaperID{10636} 
\def\confYear{CVPR 2021}

\begin{document}

\title{\ Unsupervised Discovery of the Long-Tail in Instance Segmentation Using Hierarchical Self-Supervision}

\author{Zhenzhen Weng, Mehmet Giray Ogut, Shai Limonchik, Serena Yeung \\
Stanford University\\
{\tt\small \{zzweng, giray98, shailk, syyeung\}@stanford.edu}
}
\maketitle

\begin{abstract}
Instance segmentation is an active topic in computer vision that is usually solved by using supervised learning approaches over very large datasets composed of object level masks. Obtaining such a dataset for any new domain can be very expensive and time-consuming. In addition, models trained on certain annotated categories do not generalize well to unseen objects. The goal of this paper is to propose a method that can perform unsupervised discovery of long-tail categories in instance segmentation, through learning instance embeddings of masked regions. Leveraging rich relationship and hierarchical structure between objects in the images, we propose self-supervised losses for learning mask embeddings. Trained on COCO \cite{lin2014microsoft} dataset without additional annotations of the long-tail objects, our model is able to discover novel and more fine-grained objects than the common categories in COCO. We show that the model achieves competitive quantitative results on LVIS \cite{gupta2019lvis} as compared to the supervised and partially supervised methods.
\end{abstract}

\section{Introduction}

\begin{figure}
\begin{center}
\includegraphics[width=\linewidth]{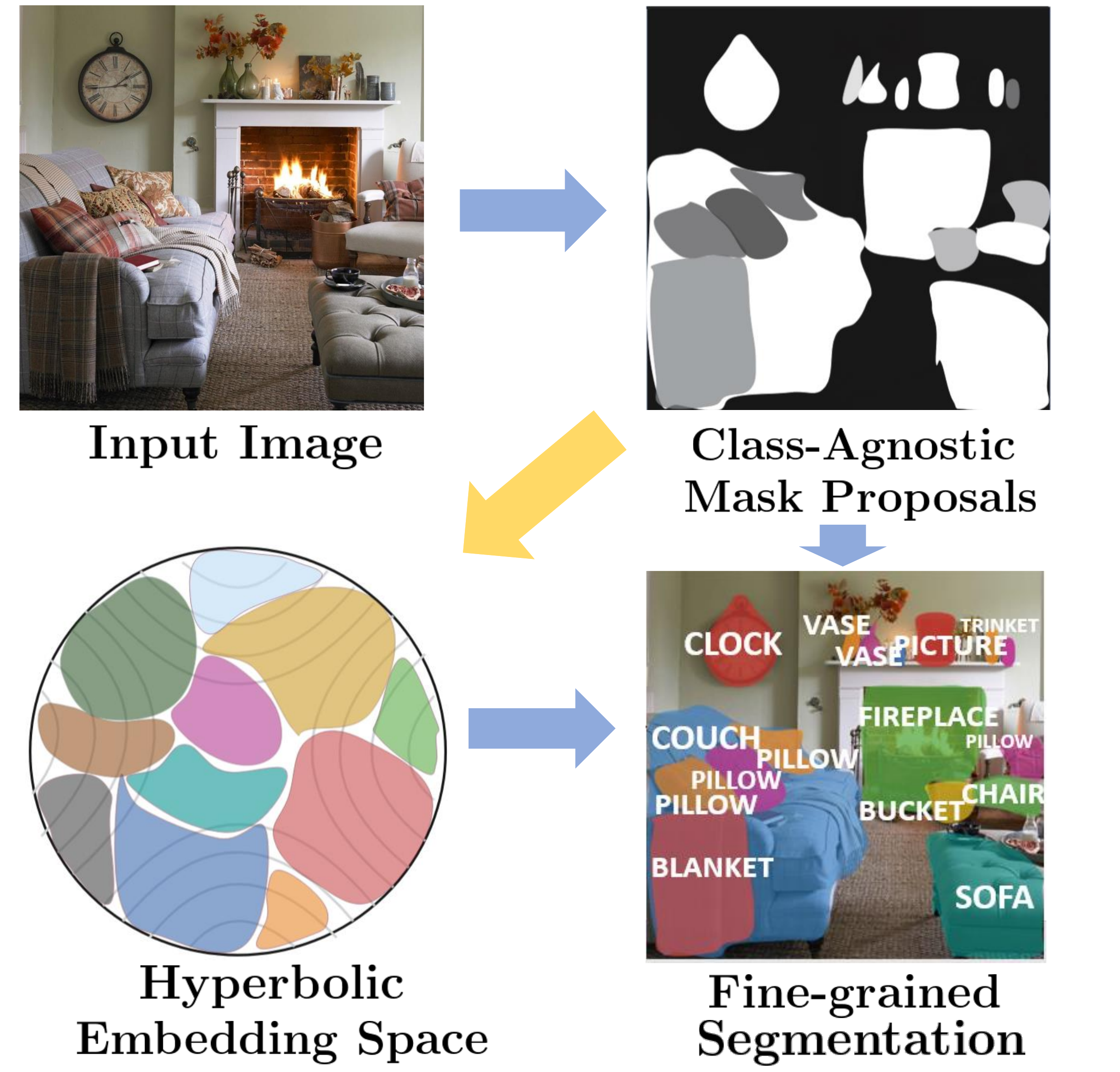}
\end{center}
   \caption{We propose an instance segmentation method that is able to discover the long-tail objects through self-supervised representation learning. Without having access to ground truth annotations of long-tail objects during training, our method is able to produce fine-grained segmentation result of novel objects. 
   }
\label{fig:general_overview}
\end{figure}

Instance segmentation is a crucial problem that has a wide range of applications in various real-world applications such as autonomous driving and medical imaging. Recent approaches \cite{hariharan2014simultaneous, he2017mask, chen2018masklab, liu2018path} have shown impressive results on large-scale datasets in parsing diverse real-world scenes into informative semantics or instance maps. Most of the existing works so far assume a fully supervised setting, where the instance level segmentation masks are available during training. However, relying on human annotated segmentation labels has a few obvious drawbacks. Since most of the existing segmentation datasets contain a small set of annotated categories (e.g. $20$ in PASCAL VOC \cite{everingham2010pascal} and $80$ in COCO \cite{lin2014microsoft}), models trained on these categories are not able to generalize well to novel and long-tail objects present in the real world. Despite the ease of capturing a large number of images nowadays, extending the set of annotated categories is still very expensive and time-consuming. Recently, a dataset for Large Vocabulary Instance Segmentation (LVIS) \cite{gupta2019lvis} was released as an attempt to increase the coverage of the annotated categories in COCO from $80$ to over $1200$ categories including the long-tail categories that appear rarely in the dataset.

To overcome the bottleneck in obtaining large annotated datasets, previous works have attempted instance segmentation with weaker forms of supervision such as the object bounding boxes \cite{dai2015boxsup, papandreou1502weakly}, points on instances \cite{bearman2016s} or image-level labels \cite{zhou2018weakly}. Recent works \cite{kuo2019shapemask, hu2018learning, fan2020commonality, zhou2020learning} aim to improve the generalization capability of instance segmentation models by employing a partially supervised setting where only a subset of classes have instance-level mask annotations during training; the remaining classes have only bounding box annotations. However, these methods still require that the bounding boxes of all categories be known prior to training, which limits their ability in discovering novel objects that do not have bounding box annotations. 

In this work we propose the first instance segmentation method that performs unsupervised discovery of the long-tail objects through representation learning using hierarchical self-supervision. To the best of our knowledge, this is the first method that eliminates the need for any type of instance level mask or bounding box annotation for the long tail of categories during training. Our idea is that since instance segmentation models (e.g. Mask R-CNN \cite{he2017mask}) trained on a small set of categories can already produce good class-agnostic mask proposals, we can leverage these masks and use representation learning to separate these proposals into distinct categories. Therefore, much of our approach, after taking the mask proposals from a region proposal network, is focused on the representation learning. We present an effective approach that exploits the inherent hierarchical visual structure of the objects and enables the embedded features of all proposed masks to be easily differentiated through unsupervised clustering. 

Our method is motivated by the recent works \cite{khrulkov2020hyperbolic, mathieu2019continuous} that use hyperbolic embeddings to boost the performance of downstream computer vision tasks. These works draw inspiration from the key observation that objects in the real world exhibit hierarchical structure. To perform self-supervised learning in hyperbolic embedding space, we introduce three triplet losses for learning better mask features and capturing hierarchical relations between the masks. We show that our model outperforms the state-of-the-art partially supervised models \cite{hu2018learning, kuo2019shapemask} even though they require box annotations of the long-tail objects during training and we do not. We also qualitatively show that our model is able to discover and segment additional novel object categories. Compared to the the fully supervised setting, our model is competitive in terms of its capability of detecting and segmenting rare and small objects.

We summarize our main contributions as the following: 
\begin{itemize}[noitemsep]
    \item We propose an instance segmentation method that is able to discover the long-tail objects using hierarchical self-supervision. To the best of our knowledge, this is the first instance segmentation method that eliminates the need for any type of ground truth annotation of the long-tail categories during training.
    \item We leverage hyperbolic embeddings to capture the hierarchical structure in the segmented objects and demonstrate the effectiveness of our learned embeddings as compared to their Euclidean counterpart.
    \item We show that our method outperforms the state-of-the-art partially supervised instance segmentation methods although using less supervision. We also provide the first set of baseline numbers on LVIS for such self-supervised methods.
\end{itemize}

\section{Related Work}
\label{related_work}
\paragraph{Fully supervised instance segmentation}
Fully supervised instance segmentation is well studied. State-of-the art methods \cite{he2017mask, chen2018masklab} use top-down approaches that first generate object bounding boxes using a region proposal network and then perform instance segmentation on each detected bounding box. On the other hand, the bottom-up approaches \cite{de2017semantic, fathi2017semantic} segment each instance directly without referring to the detection results. These approaches rely on object-level mask annotations during training, which may restrict their applicability to real world problems.

\paragraph{Partially or weakly supervised instance segmentation}
Prior works \cite{hu2018learning, kuo2019shapemask, zhou2020learning, fan2020commonality} assume a partially supervised setting where all the training images have ground truth bounding boxes, but only a subset of the images have segmentation annotations. While showing competitive segmentation results as compared to the fully supervised oracles, these works also showed great generalization ability in class-agnostic learning.

\cite{hu2018learning} takes a detection-based approach where a learned weight transfer function predicts how each class should be segmented based on parameters learned for detecting bounding boxes. To generalize to novel categories, it uses class-agnostic training which treats all categories as one foreground category. \cite{kuo2019shapemask} takes a bottom-up approach where it begins with pixel-wise representations of the objects. Assuming that novel objects share similar shapes with the labeled common objects, \cite{kuo2019shapemask} leverages aggregated shape priors as the intermediate representation to improve the generalization of the model to novel objects. \cite{zhou2020learning} starts with a saliency map for each box detection and generates a shape representation which serves as shape prior to improve segmentation quality. \cite{fan2020commonality} parses shape and appearance commonalities through instance boundary prediction and modeling pairwise affinities among pixels of feature maps to optimize the separability between instance and the background. 

Other weakly supervised instance segmentation methods use even less annotation data, such as box-level annotations \cite{khoreva2017simple, hsu2019weakly}, image-level annotations \cite{zhou2018weakly}, and image groups \cite{hsu2019deepco3}. To our knowledge, there are no prior instance segmentation methods that do not consume any kind of annotation data during training.

\paragraph{Unsupervised object discovery}
Unsupervised object discovery has long been attempted in computer vision but remains a challenging task. Earlier unsupervised methods explored partial correspondence and clustering of local features \cite{grauman2006unsupervised}, clustering by composition \cite{faktor2012clustering}. All of these approaches have been successfully demonstrated in a restricted setting with a few distinctive object classes, but their localization results turn out to be far behind weakly supervised results on challenging benchmarks. \cite{cho2015unsupervised} proposed an unsupervised approach using part-based matching and achieved state-of-the-art localization results on a challenging dataset Pascal VOC \cite{everingham2010pascal}. However, \cite{cho2015unsupervised} requires that the dataset contain multiple dominant object classes and a few noisy images without any target objects, which is not a valid assumption in datasets with complex scenes and semantics such as COCO \cite{lin2014microsoft}.

\paragraph{Self-supervised learning} Self-supervised learning constructs pretext tasks by exploiting training signals directly from the input data without external supervision. Self-supervised learning is proven to be effective in a range of pretext computer vision tasks such as patch position prediction \cite{doersch2015unsupervised, nathan2018improvements}, colorization \cite{zhang2016colorful, larsson2016learning, larsson2017colorization}, inpainting \cite{pathak2016context} and representation learning \cite{ren2018cross}. Previous works also attempted using self-supervision in instance segmentation. \cite{bewley2014online} utilized self-supervised signals from the temporal consistency of the frames in video data. \cite{pathak2018learning} gains supervision signals through the interaction between an active agent and the segmented pixels. \cite{hung2019scops} proposed a self-supervised approach for part segmentation, where they use the intuition that parts provide a good intermediate representation of objects and devise several loss functions that aids in predicting part segments that are geometrically concentrated. However, self-supervised representation learning for instance segmentation for in-the-wild images is rarely explored. In this work, we present a novel method for instance segmentation through mask-level representation learning with self-supervised triplet losses. 

\paragraph{Representation learning}
Object representations learned from large datasets can be used as intermediate representations on downstream tasks such as image classification. One popular class of methods for unsupervised representation learning uses generative models. Generative approaches \cite{hinton2006fast, kingma2013auto, goodfellow2014generative} learn to generate or model pixels in the input space. However, pixel-level generation or reconstruction is computationally expensive and may not be necessary for learning intermediate object representations. In addition, pixel-based objectives can lead to such methods being overly focused on pixel-based details, rather than more abstract latent factors. 
Unlike generative methods that focus on pixel-level generation, contrastive methods \cite{bachman2019learning, henaff2019data, chen2020simple} learn visual representations by contrasting positive and negative examples in the latent space. In our work, we learn the representations of the segmented foreground objects using a pretrained feature extractor and a hyperbolic embedding space. We sample anchor objects and the corresponding positive and negative examples and use triplet loss to learn the object level representations. In addition, we argue that the representations of the foreground region within a bounding box should be closer to the full region relative to the background region, and thus by using triplet loss in the same embedding space, we are able to augment the mask features.

\paragraph{Hyperbolic embeddings}
Most existing metric learning methods in computer vision use Euclidean or spherical distances. Natural images often exhibits hierarchical structures \cite{khrulkov2020hyperbolic, mathieu2019continuous} which makes hyperbolic embeddings a better choice than their Euclidean counterpart. In this work, we learn a hyperbolic embedding space for the mask level features. Following previous works \cite{ganea2018hyperbolic, khrulkov2020hyperbolic, nickel2017poincar}, we embed the features into a Poincaré ball model with negative curvature, which is a special instance of hyperbolic manifolds and is suitable for this particular task because of its simplicity and differentiable distance function.

\label{approach}
\begin{figure*}[tp]
\label{fig:overview}
  \begin{center}
\includegraphics[width=\textwidth]{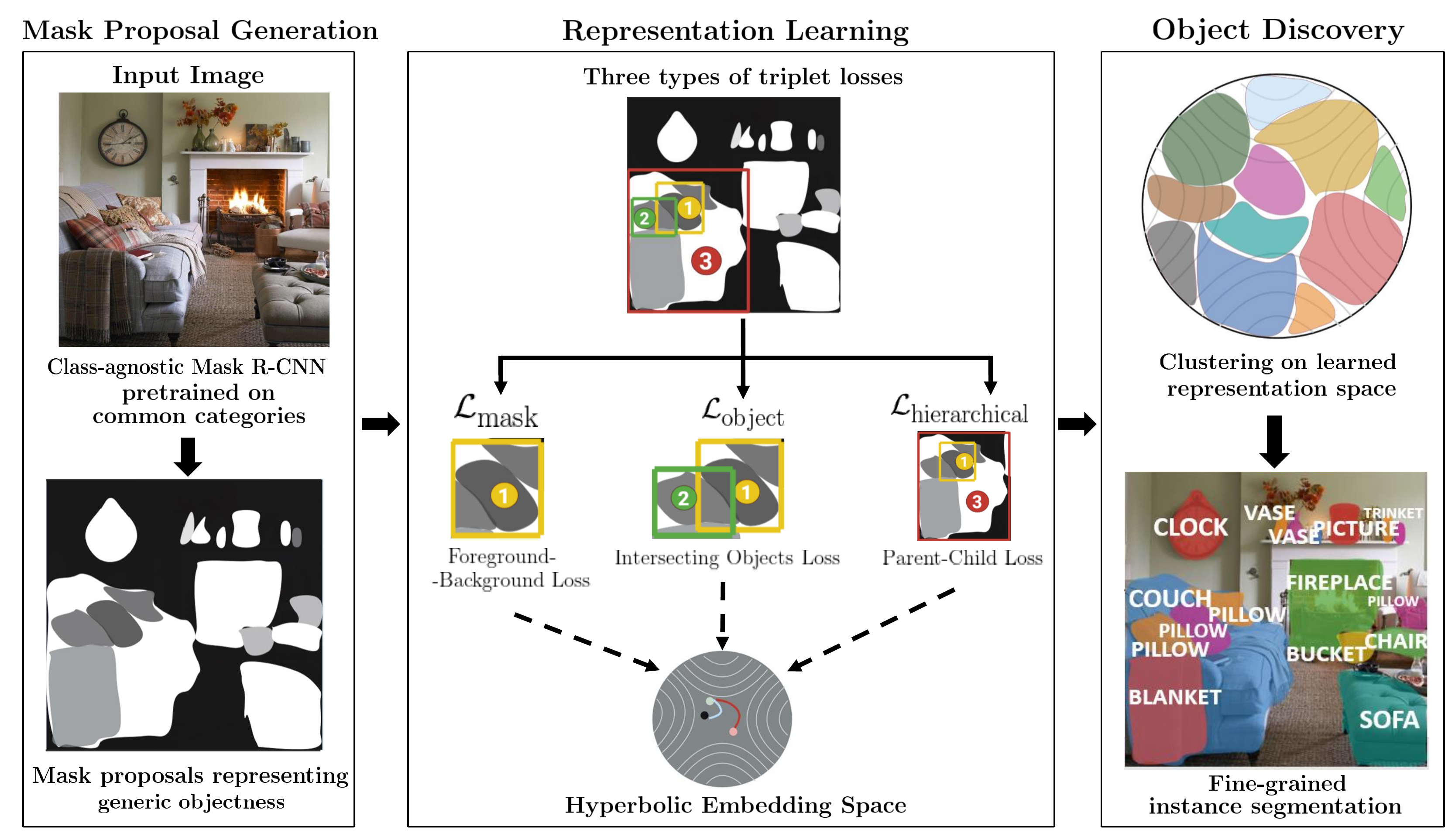}
  \end{center}
  
  \caption{Overview of our method. We first use a class-agnostic mask proposal network (pre-trained on common categories) to generate masks for all possible objects in the image (Section \ref{section:mask_proposal}). We sample the masks using sampling rules that exploits the relationship and hierarchical structure within the proposals. Then, we embed the sampled mask features into a Poincaré ball using self-supervised triplet losses (Section \ref{section:embed}). Once we have the trained Poincaré embedding space, we perform unsupervised hyperbolic clustering to identify the distinct object categories of the embedded masks (Section \ref{section:discovery}). The output of our method consists of fine-grained instance segmentation masks. }
\end{figure*}

\section{Proposed Method}
Our method is composed of three main parts: (i) mask proposal generation, (ii) representation learning with hyperbolic embeddings, and (iii) discovery of novel objects using unsupervised hyperbolic clustering. An overview of our method is shown in Figure \ref{fig:overview}. We first use a class-agnostic mask proposal generation network (pre-trained on common categories) to generate segmentation mask proposals of possible objects in the image. The goal of pre-training is for the mask proposal network learn to produce mask proposals that represent generic objectness from visual features. Although the mask proposal generation network is pre-trained with ground truth segmentation annotations of common categories, it has not been trained to detect or segment novel, long-tail categories.

Once we have the mask proposals from the segmentation head, we sample mask triplets according to the procedure in Section \ref{section:embed}. Then, we extract the mask features and embed them into a Poincaré ball using contrastive triplet losses. We use sampling mechanisms to sample the masks such that the hierarchical structure in the proposed masks is fully exploited and encoded in the hyperbolic space. In the end, once we have trained a hyperbolic embedding space, we can perform unsupervised hyperbolic clustering to identify the distinct object categories of the embedded masks.

As compared to previous partially or weakly supervised methods that are only trained to detect and segment categories that exist in the ground truth annotations, our method considers a larger set of categories by keeping all the class-agnostic mask proposals. Therefore, while previous methods such as \cite{kuo2019shapemask, hu2018learning, fan2020commonality, zhou2020learning} can only segment a pre-defined set of categories, our method is able to discover novel object categories, and we do this through effective representation learning in hyperbolic space using self-supervised signals. We will explain each part in detail in Section \ref{section:mask_proposal}, \ref{section:embed} and \ref{section:discovery}.

\subsection{Mask Proposal Generation}
\label{section:mask_proposal}
The first step of our method is to generate class-agnostic mask proposals of all possible objects in the input images. Concretely,
our mask proposal generation network takes a batch of $N$ training images $\mathcal{I} = \{I_1, I_2, ..., I_N\}$. Given an image $I \in \mathcal{I}$, we first use a mask proposal network to generate $k$ proposed bounding boxes $\mathcal{B}=\{B^1,B^2,...,B^k\}$ and the object segmentation masks within those bounding boxes $\mathcal{M}=\{M^1,M^2,...,M^k\}$. The mask proposal network is pre-trained on a large enough dataset with common categories (e.g. the 80 common categories in COCO). We refer to the segmented object within the bounding box $B^j$ as foreground $x^{j,fg}$, and the rest background $x^{j,bg}$. We then learn the relationships between the detected objects through sampling and learning a hyperbolic embedding space using the sampled mask features. This process is illustrated in Figure \ref{fig:overview}.

\subsection{Representation Learning with Hyperbolic Embeddings}
\label{section:embed}
Once we have the foreground $x^{j,fg}$ and background $x^{j,bg}$ for the proposals generated from the previous step, we sample triplets and use a feature extractor $f$ (e.g. ResNet \cite{he2016deep}) to get their representations in a hyperbolic space, $f(x^{j,*}) = z^{j,*}$ for object $j$. 

First, inspired by the intuition in \cite{song2018mask} that the dominant features of a bounding box detection should come from the visual features of the segmentation within this bounding box whereas the background of the bounding box (i.e. bounding box area that is not part of the segmentation) should contain less meaningful features, we consider a triplet loss that enables us to learn better mask features. Specifically, for each proposed segmentation mask, we encourage the mask features and the bounding box features to be close to each other. In addition, the features of the background content within this bounding box should not be as close to the mask features as the bounding box features. This is particularly important in the context of the long-tail nature of our task, since the foreground and background may often correspond to distinct objects in the visual hierarchy of the scene. The resulting triplet loss $\mathcal{L}_{\mathrm{mask}}$ is
\small
\begin{multline}
    \mathcal{L}_{\mathrm{mask}} = \sum_{j\in [k]}\max(0,
    \alpha - d(z^{j,\mathrm{full}},z^{j,\mathrm{bg}}) + d(z^{j,\mathrm{full}}, z^{j,\mathrm{fg}})) 
\end{multline}
\normalsize
where $\alpha$ is the margin of the triplet loss, and $d$ is the distance function in the respective embedding space. Formally, the distance function defined on a Poincaré ball with curvature $-1$ is defined as 
\small
\begin{equation}
d(x, y) = \cosh^{-1} (1+2\frac{||x-y||^2}{(1-||x||^2)(1-||y||^2)}) 
\end{equation}
\normalsize
We use Euclidean distance for the respective ablative experiments in the Euclidean space.

Concurrently, we sample triplets for learning the object representations. For each segmented bounding box, we randomly choose an overlapping proposal that exceeds a pre-defined IoU threshold ($0.4$) to be the positive sample, and a non-overlapping proposal to be the negative sample. The negative sample can be from either the same image, or a different image in the training batch. The idea is that if multiple proposals suggested by the region proposal network have high overlapping area, then they are likely to be the same object. Note that although Non-maximum Suppression (NMS) does get rid of high overlapped boxes, it still leaves some remaining overlapping boxes. We notice that such high IoU proposals tend to exist when there are few dominant objects in the image. In such cases, the region proposal network would tend to make proposals of different parts of the same object repeatedly. This triplet loss comes from the intuition that since those proposals are essentially of the same object, their feature embeddings should also be close to each other in the embedding space. Hence, this loss term plays a similar role as the common transformation-based loss in contrastive learning. Formally, the triplet loss $\mathcal{L}_{\mathrm{object}}$ is defined by
\small
\begin{multline}
    \mathcal{L}_{\mathrm{object}} = \sum_{j\in [k]} \max(0,
    \alpha - d(z^{j,\mathrm{fg}}, \hat{z}^{j,\mathrm{fg}}) + d(z^{j,\mathrm{fg}}, \bar{z}^{j,\mathrm{fg}})) 
\end{multline}
\normalsize
where $\hat{z}^{j,fg}$ is the feature of the positive sample, and $\bar{z}^{j,fg}$ is the feature of the negative sample.

Last but not the least, we include a hierarchical loss term which is essential in preserving the hierarchical structure among the embedded features. Specifically, for each anchor mask proposal, we find another mask that falls into the area of the anchor mask but is significantly smaller in size. As compared to the previous sampling procedure for $\mathcal{L}_{\mathrm{object}}$, the anchor mask and the positive mask proposal in the hierarchical loss would have a much smaller IoU. We refer to the anchor mask as the ``parent" and this sampled mask ``child" since the sampled mask is likely to be a sub-component of the anchor object. We use a Poincaré ball with negative curvature as the embedding space, because a Poincaré ball is able to capture this type of hierarchical parent-child relations within the data with arbitrarily low distortion as compared the Euclidean space. Therefore, we utilize this useful property of Poincaré ball and explicitly enforce that the mask features of the ``parent" object to be closer to the origin of the tree than the mask features of the ``child" object. Formally, we include a loss term $\mathcal{L}_{\mathrm{hierarchical}}$,
\small
\begin{multline}
    \mathcal{L}_{\mathrm{hierarchical}} = \sum_{j\in [k]}\max(0,
    \alpha - d(z^{j,\mathrm{child}}, o) + d(z^{j,\mathrm{fg}}, o)) 
\end{multline}
\normalsize
where $o$ represents the origin of the Poincaré ball, and $z^{j,\mathrm{child}}$ is the feature of the child mask of proposal $j$.

The total loss for image $I$ is a weighted sum of three loss terms.
\begin{equation}
\label{eq:loss}
    \mathcal{L} = \beta \mathcal{L}_{\mathrm{mask}} + \gamma \mathcal{L}_{\mathrm{object}} + \mathcal{L}_{\mathrm{hierarchical}}
\end{equation}

\subsection{Discovery of Novel Objects}
\label{section:discovery}
We validate our approach using LVIS \cite{gupta2019lvis} which contains ground truth annotations for the long-tail categories in the COCO images. In order to evaluate our approach against existing fully supervised or partially supervised benchmarks as well as showing that our instance embeddings have learned diverse object representations, we perform hyperbolic K-means clustering (Figure \ref{fig:overview}) with $K$ being a much larger number than the number of categories in the annotations. We use a large $K$ in order to account for the large number of long-tail or novel categories. We pick a few ground truth masks from each category and use those embedded ground truths as anchors in assigning labels to our detected clusters. 

In our evaluation, we look at the clustered mask features and assign a label to each cluster by taking the label of the closest ground truth category. Once a cluster that is on average closest to a ground truth cluster has been assigned the corresponding label, other clusters are no longer eligible to be assigned to that label. This more rigorously considers our clusters to be good only if they have both high precision as well as recall, since the same label concept should not be split across multiple clusters if our goal is category discovery. With this process, clusters not assigned to any LVIS labels are considered “novel”.

In the COCO to LVIS experiment in Section \ref{experiments}, we show that with a mask proposal network pretrained on COCO categories, our method is effective in discovering long-tail objects that are not among the common categories in COCO. 

\begin{table*}[th]
\small
  \centering
  \scalebox{0.9}{%
  \begin{tabular}{ll|lll|lll|lll}
    \toprule
    Model & Supervision & mAP & mAP$_{50}$ & mAP$_{75}$& mAP$_{r}$ & mAP$_{c}$ & mAP$_{f}$& mAP$_{s}$ & mAP$_{m}$ & mAP$_{l}$ \\
    \midrule
    Mask R-CNN&Fully Supervised&0.201&0.327& 0.212&0.072&0.199&0.284&0.106&0.214&0.325 \\
    \midrule
    \midrule
    ShapeMask \cite{kuo2019shapemask} & COCO masks+LVIS boxes & 0.084 & 0.137 & 0.089 &0.056 & 0.084 & 0.102 & 0.062 & 0.088 & 0.103  \\
    \midrule
    Mask$^X$ R-CNN \cite{hu2018learning} & COCO masks+LVIS boxes & 0.056 &0.095&0.058 & 0.024 & 0.051 & 0.079 & 0.031 & 0.056 & 0.078  \\
    \midrule
    Ours (rand. init. backbone) & COCO masks & 0.096 & 0.139 & 0.104 & 0.051 & 0.092 & 0.168 & 0.075 & 0.107 & 0.139   \\
    \midrule
    \textbf{Ours} & \textbf{COCO masks} & \textbf{0.109} & \textbf{0.160} & \textbf{0.113}&\textbf{0.087}&\textbf{0.105}&\textbf{0.174}&\textbf{0.092}&\textbf{0.129}&\textbf{0.147}  \\
    \bottomrule
  \end{tabular}
  }
  \vspace*{2mm}
  \caption{Quantitative results of the COCO to LVIS generalization experiment. We report mAP for the small/medium/large objects and the rare/common/frequent objects as defined by LVIS. Our ResNet-50 for mask proposals was pretrained on ImageNet~\cite{russakovsky2015imagenet}; a version with random initialization ``Ours (rand. init. backbone)" instead drops performance slightly from $0.109$ to $0.096$ overall mAP. However, the crippled model still outperforms the prior work baselines, despite the fact that ShapeMask and Mask$^X$ R-CNN do use backbones pretrained on ImageNet in addition to their extra utilization of bounding boxes of the LVIS categories.}
  \vspace*{-1mm}
 \label{table:baselines}
\end{table*}

\section{Experiments}
\label{experiments}
\paragraph{Datasets} 
We pre-train our class-agnostic mask proposal network on COCO~\cite{lin2014microsoft} that contains $80$ common categories. We believe the true strength of our method lies in its ability to find new objects besides the $80$ common object classes. Therefore, we validate our approach using LVIS \cite{gupta2019lvis} which contains ground truth annotations for the long-tail categories in the COCO images. LVIS contains ground truth instance-level segmentations for over $1200$ categories including sparsely populated categories that appear fewer than $5$ times in the training set. All the $80$ common categories in COCO can be mapped to LVIS. 

\paragraph{Metrics}
Following previous works we use mean average precision (mAP) for quantitative evaluation. We split the LVIS categories into $2$ subsets: the $80$ common categories that exist in COCO annotation and the novel (long-tail) categories that are not in COCO. We report mAPs on the second subset (i.e. novel categories). To show that our method generalizes well to long-tail categories, we perform qualitative examination and cluster analysis on the discovered novel categories.

\subsection{Implementation Details}
\label{experiments:RPN}
Our mask proposal network is built upon the class-agnostic version of Mask R-CNN \cite{he2017mask}. Specifically, all the bounding box proposals (after NMS) are passed to the class-agnostic mask branch directly without going through the classification branch. The mask proposal network is pre-trained with the $80$ common categories in COCO. We use ResNet-50 \cite{he2016deep} as the backbone network. Note that the pre-training is only used for the model to learn to propose and segment all objects representing general objectness using visual features. We do not inject any information about the novel objects or the category information of the common objects during the pre-training stage. Since our mask proposal method is class-agnostic in the sense that we keep all the proposals regardless of the classes they might be in, we are not discarding any novel or long-tail categories as previous supervised instance segmentation methods do.

For each image, we keep $50$ object proposals with the highest objectness score after using non-maximum suppression (NMS) with threshold $0.75$. When calculating the triplet losses, we sample $1$ positive example and $3$ negative examples for each anchor according to the procedure described in Section \ref{section:embed}. Therefore, for each anchor mask, we calculate $3$ triplet loss terms. Once we have the triplet samples, we use a randomly initialized ResNet-18 followed by $2$ fully-connected layers of size $256$ and $64$ and finally an exponential map to map the features to the Poincaré ball. We use a $2$-dimensional Poincaré ball with curvature $-1$, and Adam \cite{kingma2014adam} optimizer with learning rate $10^{-4}$. The weight $\beta$ and $\gamma$ in the loss function (Equation \ref{eq:loss}) are both $0.2$. We use hyperbolic K-Means with k=$1500$. We will justify our selection of these hyper-parameters in Section \ref{sec:ablations}.
\begin{figure*}
 \begin{center}
     \includegraphics[width=\linewidth]{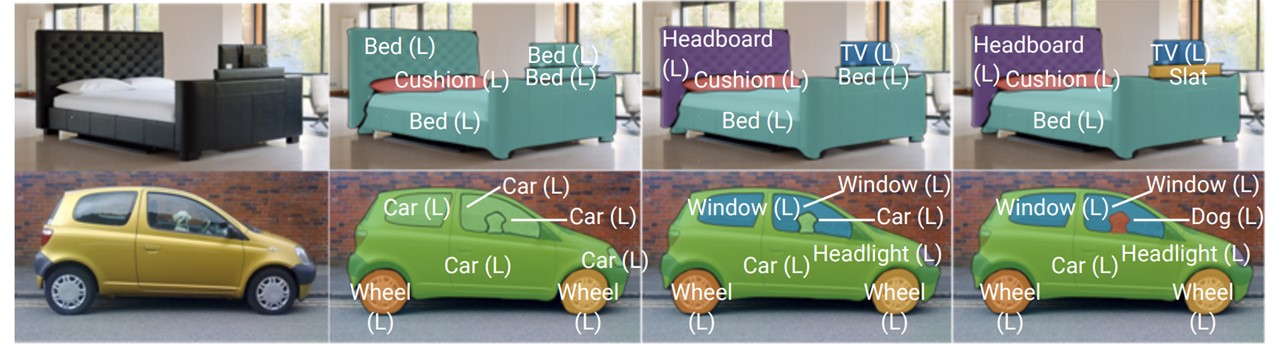}
 \end{center}
 \caption{Qualitative results showing model ablations. \textbf{From left to right}: Original image; segmentation masks obtained using only mask loss term; with mask loss and object loss; with all three loss terms. ``L" means LVIS categories; w/o ``L" means additional novel categories. We show that the hierarchical loss is essential for the model to discern the sub-components of an object (e.g. the slat and the bed). }
 \label{fig:ablation}
\end{figure*}
\subsection{Comparison with Partially Supervised Baselines}

We compare our model with ShapeMask \cite{kuo2019shapemask} and Mask$^X$ R-CNN \cite{hu2018learning} given the similar focus of segmenting novel objects. Both baselines consider a partially supervised setting, where the mask annotations for only a small subset of the categories and the box labels for all the categories are available during training. In the context of COCO to LVIS experiment, \cite{kuo2019shapemask} and \cite{hu2018learning} assume that bounding boxes for all categories in LVIS are available, but masks for only the categories in COCO are available during training. However, we do not assume the availability of the bounding boxes of the LVIS categories. In addition to \cite{kuo2019shapemask} and \cite{hu2018learning}, we also compare our model with the fully supervised Mask R-CNN with the same backbone architecture. The quantitative results are reported in Table \ref{table:baselines}. Our method outperforms both \cite{hu2018learning} and \cite{kuo2019shapemask} although we use less supervision. On the rare and small object categories, our model is comparable to the fully supervised Mask R-CNN model. The discrepancy of performance across different types of object categories is small compared to the fully supervised model, which can perform much worse on rare objects than on frequent categories.

\subsection{Ablation Studies}
\label{sec:ablations}
\paragraph{Effect of the Number of Region Proposals}
Besides concerns of computational cost, the number of mask proposals to keep per image could influence the performance of our model. Keeping too few proposals would limit our ability to discover novel objects, and keeping too many proposals would introduce too much noise to the embeddings and worsen our cluster quality. By varying the number of region proposals (Table \ref{table:ablation}) we observe that increasing the number of proposals per image leads to better segmentation results. However, as the number of proposals gets too big, the improvement becomes marginal.

\begin{table} 
\small
  \centering
  \scalebox{0.9}{%
  \begin{tabular}{l|lll}
    \toprule
    Model & mAP & mAP$_{50}$ & mAP$_{75}$ \\
    \midrule
    No. RP = 20 & 0.0318 &  0.0459  & 0.0332 \\
    No. RP = 50 & 0.1086 & 0.1597 & 0.1125  \\
    No. RP = 100  & 0.1083 & 0.1643 & 0.1117  \\
    \midrule
    $\alpha$ = 0.1 & 0.0806 & 0.8680 & 0.0844 \\
    $\alpha$ = 0.2 & 0.1086 & 0.1597 & 0.1125 \\
    $\alpha$ = 0.5 & 0.0987 & 0.1455 & 0.0982  \\
    \midrule
    $\beta$ = 0.1 & 0.0526 & 0.6280 & 0.0544 \\
    $\beta$ = 0.2 & 0.1086 & 0.1597 & 0.1125 \\
    $\beta$ = 0.5 & 0.0614 & 0.7030 & 0.0639  \\
    \midrule
    w/o $\mathcal{L}_{\mathrm{mask}}$ & 0.0689 &  0.0842  & 0.0707 \\
    w/o $\mathcal{L}_{\mathrm{object}}$ & 0.0374 & 0.0455 & 0.0396   \\
    w/o $\mathcal{L}_{\mathrm{hierarchical}}$ & 0.0846 & 0.1082 & 0.0921  \\
    \midrule
    Euclidean & 0.0382 & 0.0641 & 0.0433 \\
    Poincaré & 0.1086 & 0.1597 & 0.1125  \\
    \bottomrule
  \end{tabular}
  }
\normalsize
  \vspace*{2mm}
  \caption{Ablation Studies. We vary the number of region proposals (No. RP), margin of the triplet losses ($\alpha$), weight of the mask loss term ($\beta$), and report the mean APs. In addition, we also report mAPs after taking out individual loss terms, as well as using an Euclidean embedding space instead of hyperbolic space.}
  \vspace*{-2mm}
  \label{table:ablation}
\end{table}
\paragraph{Effect of the Margin}
Our representation learning process requires sampling of triplets. However, the number of possible triplets grows cubically as we increase the batch size. Furthermore, choosing hard negatives for tail objects is already difficult given their scarce number whereas it is much easier to find negative samples for common objects. In order to account for this heterogeneity in the possible hard negative samples, we instead vary the margin parameter ($\alpha$) of the loss function since the margin parameter intrinsically represents the sampling effect. We observed from Table \ref{table:ablation} increasing $\alpha$ indeed improves the performance of our model. However, after around the value of $0.2$, increasing the margin seem to have minimal effect.
\vspace*{-2mm}
\paragraph{Effect of the Loss Terms}
Mask loss term in our loss function serves the important role of distinguishing between background and foreground objects. However, we wanted to be sure that the remaining losses are also important for clustering and that their contribution is no less important. Therefore, we gradually increased the weight of the mask loss term and noted that although a very low mask loss term results in worse background segmentation and hence distorted object embeddings, increasing the mask loss does not lead to a monotonic increase in the mean average precision. Overall, our loss function has three terms, and we show that each term is crucial for the performance of the model by taking out individual loss term at a time (Table \ref{table:ablation}). In Figure~\ref{fig:ablation} we include qualitative results of the ablated loss function, showing that the three loss terms are essential in learning good representations.

\begin{figure*} 
\begin{center}
\includegraphics[width=0.87\linewidth]{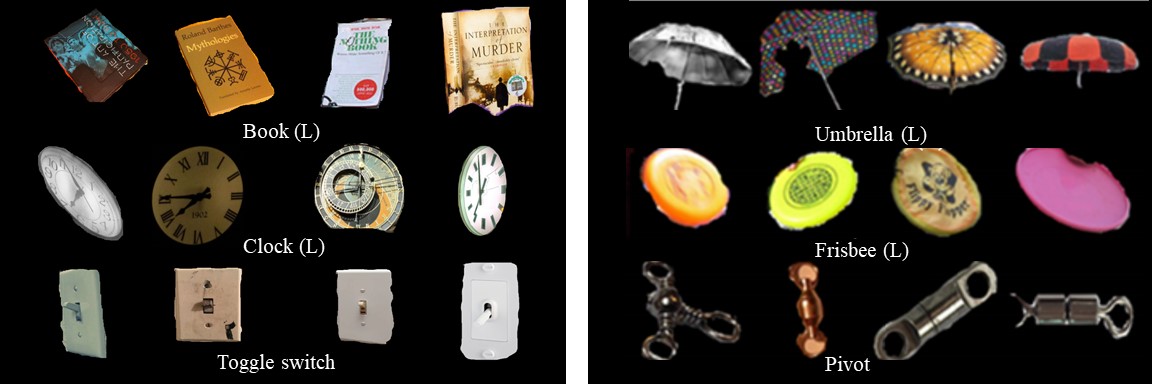}
\end{center}
\vspace*{1mm}
\caption{Examples of the segmentation masks of the discovered categories. ``L" means LVIS categories; without ``L" means additional novel categories. Left sub-figure: Examples from ``clock" and ``book" categories in LVIS, as well as the ``toggle switch" category that is not in LVIS, but discovered by our model. Right sub-figure: Examples from ``umbrella" and ``frisbee" categories in LVIS, as well as the ``pivot" category that is not in LVIS, but discovered by our model.
}
\label{fig:new_object_clusters}
\end{figure*}

\vspace*{-2mm}
\paragraph{Euclidean vs. Hyperbolic Space}
We opt for a hyperbolic embedding space because of its capability of embedding hierarchical data with arbitrarily low distortion. To validate our hypothesis of the suitability of hyperbolic space over its Euclidean counterpart, we compare our model's performance with Euclidean versus Poincaré embedding space. In Table \ref{table:ablation}, we show that the Poincaré embedding space brings a massive improvement over the Euclidean embedding space in terms of mean average precision.
\vspace*{-2mm}
\paragraph{Cluster Analysis}
After embedding mask features into the hyperbolic space, we use hyperbolic K-Means clustering to group the mask proposals. We use elbow method \cite{thorndike1953belongs} to pick the number of clusters. Elbow method is a heuristic used for determining the optimal number of clusters using the explained variation as a function of the number of clusters. In addition, we use purity scores to measure the quality of the resulting clusters. In the context of cluster analysis, purity score is a measure of the extent to which clusters contain a single class \cite{schutze2008introduction} and is a commonly-used criterion of cluster quality. In Table \ref{table:purity}, we show purity scores resulting from different number of clusters, including the optimal cluster number determined by elbow method. Table \ref{table:purity} suggests that the objects within the discovered clusters have strong semantic coherence, particularly when the clustering is done using the number of clusters chosen by elbow method.
\begin{table}[h]
  \small
  \centering
  \scalebox{0.86}{%
  \begin{tabular}{l|l|llll}
    \toprule
    No. of Clusters & LVIS & Purity$_{Avg}$ & Purity$_{r}$ & Purity$_{c}$ &  Purity$_{f}$ \\
    \midrule
    k=1300 & 982 &0.534 & 0.335 & 0.574 & 0.615\\
    \midrule
    k*=1462 (Elbow) & 1079 &0.582 & 0.387 & 0.667 & 0.688\\
    \midrule
    k=1500 & 1083 & 0.551 & 0.346 & 0.652 & 0.712\\
    \midrule
    k=2000 & 1106 &0.419 & 0.218 & 0.660 & 0.723\\
    \bottomrule
  \end{tabular}
  }
  \normalsize
  \vspace*{1mm}
  \caption{Cluster purity analysis with different number of clusters. The ``LVIS" column includes the number of discovered clusters determined to correspond to LVIS classes (and used to compute the purity scores).}
  \vspace*{-3mm}
  \label{table:purity}
\end{table}
\subsection{Qualitative Examples of New Object Discovery}
After obtaining clusters using hyperbolic K-means, we match these clusters with the nearest cluster obtained from ground-truth LVIS annotations. This helped us see how successful we were in discovering the long-tail in the LVIS. We then mapped a few embeddings in various clusters to their original segmentation mask to get a sense of whether these objects were correctly grouped. We include qualitative examples of the discovered mask clusters in Figure \ref{fig:new_object_clusters}. As shown, we are able to discover LVIS categories such as ``clock", ``book" and ``frisbee". Beyond discovering LVIS categories that are not in COCO, we also found additional object classes that do not correspond to any LVIS categories (e.g. ``toggle switch" and ``pivot" in Figure~\ref{fig:new_object_clusters}). Since these additional object classes are not annotated in LVIS, they are therefore impossible to be found by any previous fully supervised or weakly/partially supervised methods.
\vspace*{-2mm}
\section{Conclusion}
In this paper we proposed an instance segmentation method that discovers long-tail objects through self-supervised representation learning. Trained on the common categories in COCO without consuming any annotations on the long-tail categories, we show that our model outperforms partially supervised methods that use box annotations of the long-tail objects. More importantly, through learning representations of more granular objects and the rich relationships between related regions, we demonstrated that our model is able to discover long-tail categories in LVIS and novel categories that are not in COCO nor LVIS.
\vspace*{-2mm}

\paragraph{Acknowledgements}
This material is based upon work supported by the National Science Foundation under Grant No. 2026498, as well as a seed grant from the Institute for Human-Centered Artificial Intelligence (HAI) at Stanford University.

\section*{Appendix}

\section*{A. Additional Experiment on VOC to non-VOC Generalization}
To show that our model is not biased towards the long-tail categories in LVIS, we conduct additional quantitative experiment for PASCAL VOC \cite{everingham2010pascal} to non-VOC generalization. In this experiment, we show that our model is able to discover the non-VOC categories in COCO even though the class-agnostic region proposal network was pre-trained on only the categories in VOC. Since the $80$ categories in COCO \cite{lin2014microsoft} also include the $20$ categories in PASCAL VOC, we pre-train our mask proposal network on the $20$ categories in PASCAL VOC, and show that our method is able to discover the objects that belong to the rest of the $60$ categories in COCO.

\paragraph{Quantitative Results}
In Table \ref{table:PASCAL_VOC_baselines} we compare the mean average precision (mAP) of the fully-supervised Mask R-CNN model, the partially-supervised methods (ShapeMask and Mask$^X$ R-CNN) and our method. Although our method uses less amount of supervision than partially-supervised methods, we are able to outperform both methods. What is more, the improvement over semi-supervised models is even stronger given the fact that detecting $60$ non-VOC categories in COCO is an easier task compared to the $1200$ long-tail categories in LVIS. Our model overall demonstrates performance comparable to that of the fully-supervised Mask R-CNN model.

\begin{figure*}[tp]
\label{fig:supplement1}
  \centering
  \scalebox{0.75}{
  \includegraphics[width=\textwidth]{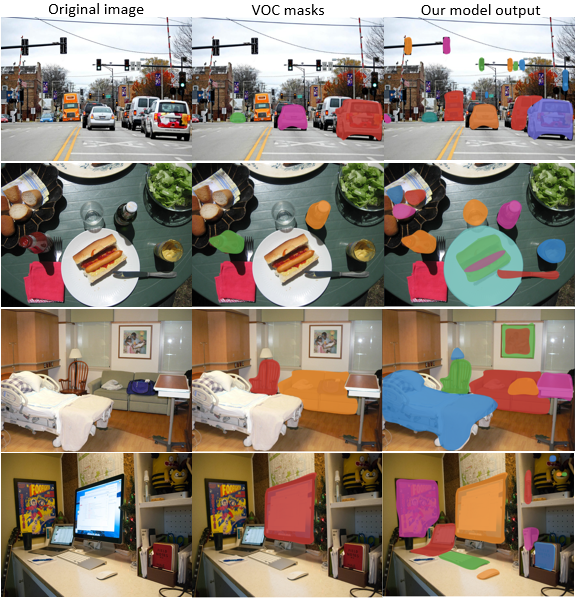}}
  \caption{Qualitative examples of new object discovery. The region proposal network was pre-trained with VOC categories in the COCO dataset. Each column shows the segmentation results on an image in the COCO dataset. }
\label{fig:quali}
\end{figure*}

\paragraph{Cluster Analysis} Following the same protocol of the ``COCO to LVIS" experiments in the main paper, we try different number of clusters in the hyperbolic clustering algorithm, and then report cluster purity scores on the final clusters (Table \ref{table:purity2}). The number of clusters that are matched to original COCO categories (excluding the $20$ Pascal VOC classes) increases as we increase the number of clusters $k$. The highest purity scores are obtained at the optimal $k$ determined by the elbow method
.
\paragraph{Qualitative Results}
In Figure \ref{fig:quali} we show qualitative examples of the new categories (i.e. non-VOC categories in COCO) discovered using our method. Each row shows the segmentation results on an image in the COCO dataset. In the first example, our model is able to segment non-VOC categories, such as traffic lights, trucks and stop signs. In the second example, our model is able to discover and segment novel categories such as glasses, knives, plates and even hot dogs and slices of bread. In the third example, our model successfully finds new classes such as bed, bag, lamp and paintings. Notice how the frame of the painting is separately detected from its canvas. In the fourth example, television is a PASCAL category. We observe that the poster, laptop, mouse, keyboard, essence bottle books as well as eyes of teddy bears could also be segmented using our method. 

\begin{table}[H]
  \centering
  \scalebox{0.75}{%
  \begin{tabular}{l|l|llll}
    \toprule
    No. of Clusters  & COCO & $\mathrm{Purity}_{\mathrm{Avg}}$ & $\mathrm{Purity}_{\mathrm{s}}$ & $\mathrm{Purity}_{\mathrm{m}}$ & $\mathrm{Purity}_{\mathrm{l}}$\\
    \midrule
    k=80  & 41  & 0.483 & 0.413 & 0.478 & 0.652\\
     \midrule
    k=90  & 44 & 0.532 & 0.468 & 0.553 & 0.681 \\
    \midrule
    k*=108 (Elbow)  & 51  & 0.622 & 0.524 & 0.637 & 0.744\\
    \midrule
    k=200  & 60  & 0.520 & 0.436 & 0.535 & 0.669\\
    \bottomrule
  \end{tabular}
  }
  \vspace*{1mm}
  \caption{Cluster purity analysis with different number of clusters. The number of discovered clusters determined to correspond to COCO classes excluding PASCAL VOC classes (and used to compute the purity scores) are denoted under the COCO column. }
  \label{table:purity2}
\end{table}

\begin{figure*}[h]
\label{fig:supplement1}
  \centering
  \scalebox{0.8}{
  \includegraphics[width=\textwidth]{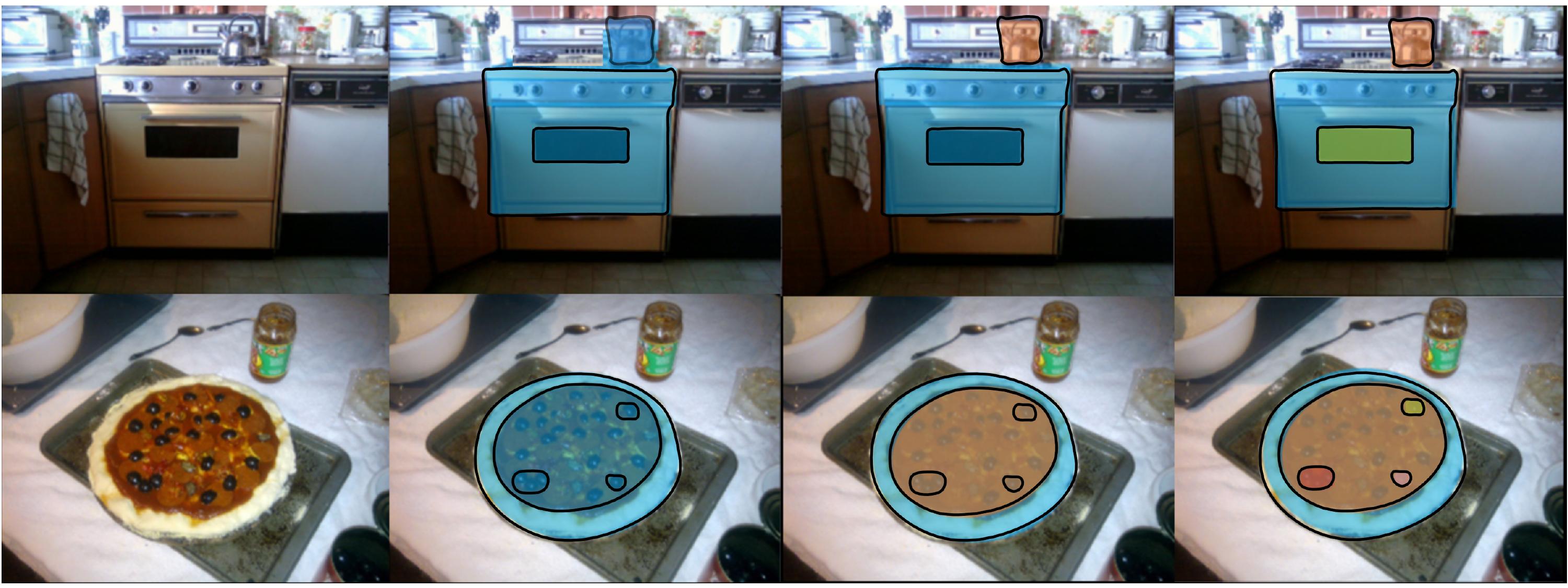}}
  \caption{Additional qualitative example showing model ablations. \textbf{From left to right}: Original image; segmentation masks obtained using only mask loss term; with mask loss and object loss; with all three loss terms included.
   }
\label{fig:additional_ablation}
\end{figure*}

\begin{table*}[h]
  \centering
  \vspace*{5mm}
  \scalebox{1}{%
  \begin{tabular}{ll|lll|lll}
    \toprule
    Model & Supervision & $\mathrm{mAP}$ & $\mathrm{mAP}_{50}$ & $\mathrm{mAP}_{75}$& $\mathrm{mAP}_{s}$ & $\mathrm{mAP}_{m}$ & $\mathrm{mAP}_{l}$ \\
    \midrule
    Mask R-CNN \cite{he2017mask} & Full supervision &0.344&0.552& 0.363&0.186&0.391&0.479 \\
    \midrule
    \midrule
    ShapeMask \cite{kuo2019shapemask} & VOC Masks + non-VOC Boxes & 0.302 & 0.493 & 0.315&\textbf{0.161}&0.382&0.384\\
    \midrule
    Mask$^X$ R-CNN \cite{hu2018learning} & VOC Masks + non-VOC Boxes & 0.238&0.429&0.235&0.127&0.281&0.335  \\
    \midrule
    \textbf{Ours} & \textbf{VOC Masks} & \textbf{0.327} & \textbf{0.525} & \textbf{0.331}&0.159&\textbf{0.385}&\textbf{0.413}\\
    \bottomrule
  \end{tabular}
  }
  \vspace*{3mm}
  \caption{Quantitative results on VOC to non-VOC. The fully-supervised Mask R-CNN is trained with the masks and boxes for all categories in COCO (i.e. including VOC and non-VOC categories). The partially-supervised methods (ShapeMask and Mask$^X$ R-CNN) are trained using the masks of the categories that are in VOC and the bounding boxes of the categories that are not in VOC. Our model consumes only VOC masks in pre-training the region proposal network. Each model was evaluated on the non-VOC categories. Our method outperforms the partially-supervised methods in terms of mAP.}
\label{table:PASCAL_VOC_baselines}
\end{table*}

\section*{B. Additional Qualitative Examples}
In Figure \ref{fig:additional_ablation} we show additional qualitative examples of model ablations. We show that the designed loss terms are essential in discovering and segmenting fine-grained objects.

\newpage

{\small
\bibliographystyle{ieee_fullname}
\bibliography{egbib}
}

\end{document}